\documentclass[conference]{IEEEtran}
\IEEEoverridecommandlockouts
\usepackage{cite}
\usepackage{amsmath,amssymb,amsfonts}
\usepackage{graphicx}      
\usepackage{multirow}      
\usepackage{makecell}      
\usepackage{pifont}        
\usepackage{rotating}      
\usepackage[table]{xcolor} 
\usepackage{url}
\usepackage{array}
\usepackage{algorithmic}
\usepackage{graphicx}
\usepackage{multirow}
\usepackage{textcomp}
\usepackage{xcolor}
\usepackage{comment}
\usepackage{pifont}
\newcommand{\cmark}{\ding{51}}%
\newcommand{\xmark}{\ding{55}}%
\usepackage{makecell} 
\def\BibTeX{{\rm B\kern-.05em{\sc i\kern-.025em b}\kern-.08em
    T\kern-.1667em\lower.7ex\hbox{E}\kern-.125emX}}
\begin{document}

\title{Advancing Autonomous Racing: A Comprehensive Survey of the RoboRacer (F1TENTH) Platform\\
}

\author{\IEEEauthorblockN{Israel Charles, Hossein Maghsoumi, Yaser Fallah}
\IEEEauthorblockA{\textit{Department of Electrical and Computer Engineering} \\
\textit{University of Central Florida}\\
Orlando, USA \\
israel.charles@ucf.edu, hossein.maghsoumi@ucf.edu, yaser.fallah@ucf.edu}
}

\maketitle

\IEEEpubid{%
  \begin{minipage}{\textwidth}
    \vspace{6.8\baselineskip}   
    \centering
    \fbox{%
      \parbox{0.92\textwidth}{\centering\small
        This paper has been accepted and presented at the AIRC 2025 conference. The final version will appear in the IEEE Xplore digital library.}%
    }%
  \end{minipage}%
}%
\IEEEpubidadjcol

\begin{abstract}
The RoboRacer (F1TENTH) platform has emerged as a leading testbed for advancing autonomous driving research, offering a scalable, cost-effective, and community-driven environment for experimentation. This paper presents a comprehensive survey of the platform, analyzing its modular hardware and software architecture, diverse research applications, and role in autonomous systems education. We examine critical aspects such as bridging the simulation-to-reality (Sim2Real) gap, integration with simulation environments, and the availability of standardized datasets and benchmarks. Furthermore, the survey highlights advancements in perception, planning, and control algorithms, as well as insights from global competitions and collaborative research efforts. By consolidating these contributions, this study positions RoboRacer as a versatile framework for accelerating innovation and bridging the gap between theoretical research and real-world deployment. The findings underscore the platform’s significance in driving forward developments in autonomous racing and robotics.
\end{abstract}


\section{Introduction}
Autonomous driving technology has advanced remarkably in recent years, driving progress in artificial intelligence, robotics, and control systems. A key factor in this advancement is the ability to rigorously test and evaluate control algorithms, sensing technologies, and real-time decision-making in dynamic yet controlled environments. Platforms designed for this purpose play a vital role in bridging the gap between theoretical development and practical deployment. Among these, the F1TENTH platform—recently rebranded as RoboRacer—has emerged as a leading tool for advancing autonomous racing research. By providing a realistic, scalable, and cost-effective testbed, it facilitates cutting-edge research, education, and innovation in this rapidly evolving field.

The RoboRacer (F1TENTH) platform, a typically 1/10th-scale autonomous vehicle system, replicates real-world challenges such as high-speed driving, trajectory planning, and real-time control. Its modular, open-source design has made it a widely adopted tool in both academic research and competitive environments. By pushing autonomous systems to their operational limits, the platform fosters advancements in precise perception, efficient path planning, and robust control strategies, offering a unique and dynamic testing environment.

This paper presents a comprehensive review of the RoboRacer platform, highlighting its significant contributions to advancing autonomous racing research. It examines the platform's core components, including its modular hardware and software design, and their unique role in enabling high-speed, dynamic scenario testing. The review also explores the use of simulation tools to support scalable and safe algorithm development and addresses the critical Sim2Real challenges in translating simulation insights into real-world performance. Additionally, the paper discusses the availability of datasets, both real-world and simulated, that aid in training, validating, and benchmarking autonomous systems. Key advancements in control algorithms, ranging from classical methods to modern learning-based approaches, are analyzed in the context of high-speed racing applications. Lastly, the survey reflects on the impact of RoboRacer competitions in shaping the field of autonomous systems research.

Through this structured review, the paper aims to provide a foundational understanding of the RoboRacer platform, address open challenges, and offer a forward-looking perspective on its potential to shape the future of autonomous racing and robotics.

\section{Background}
\subsection{Importance of Autonomous Racing}
Autonomous racing, particularly through the RoboRacer platform, plays a pivotal role in advancing broader autonomous driving technologies. Racing environments are ideal for testing control systems and perception algorithms due to the time-critical nature of these tasks, the necessity for rapid decision-making, and the inherent complexity of the environment. In contrast to traditional self-driving research that often focuses on urban or highway scenarios \cite{10137425}, autonomous racing introduces challenges involving real-time trajectory optimization, management of high-speed dynamics, and maintaining vehicle stability at the threshold of physical limits \cite{roboracepaper1}, \cite{roboracepaper2}.

\renewcommand\theadfont{\bfseries}

\begin{table*}[t]
\centering
\caption{Detailed Comparison of Autonomous Racing Platforms}
\resizebox{\textwidth}{!}{%
\begin{tabular}{|l|c|c|c|l|l|l|c|c|l|c|c|}
\hline

\thead{Platform} & 
\thead{Scale} & 
\thead{Cost} & 
\thead{OS} & 
\thead{Target Audience} & 
\thead{Primary Use Cases} & 
\thead{Sensor Suite} & 
\thead{Performance} & 
\thead{Comm. Supp.} & 
\thead{HW Flex.} & 
\thead{Act.} & 
\thead{Sim.} \\

\hline
\hline
\makecell{Indy Autonomous \\ Challenge (IAC) \cite{indyautonomouspaper1}} & \makecell{FS \\ } & \makecell{High \\ } & \makecell{\xmark \\ } & \makecell{Academic, \\ Industry \\ } & \makecell{High-performance \\ autonomous racing} & \makecell{LiDAR, GNSS/INS \\ Cameras} & \makecell{High- \\ Speed} & \makecell{Moderate \\ } & \makecell{Proprietary \\ } & \makecell{\cmark \\ } & \makecell{\cmark \\ } \\
\hline
\makecell{Roborace \cite{roboracepaper1}\\ } & \makecell{FS \\ } & \makecell{High \\ } & \makecell{\xmark \\ } & \makecell{Industry, \\ Academic} & \makecell{High-performance \\ autonomous racing} & \makecell{Cameras, \\ LiDAR, GPS} & \makecell{High- \\ Speed} & \makecell{Limited \\ } & \makecell{Proprietary \\ } & \makecell{\xmark \\ } & \makecell{\cmark \\ } \\
\hline
\makecell{Audi Autonomous \\ Driving Cup \cite{audi}} & \makecell{SS \\ } & \makecell{Med. \\ } & \makecell{\xmark \\ } & \makecell{Academic \\ } & \makecell{Algorithm \\ development} & \makecell{Cameras, \\ LiDAR} & \makecell{Low- \\ Speed} & \makecell{Limited \\ } & \makecell{Proprietary \\ } & \makecell{\cmark \\ } & \makecell{\xmark \\ } \\
\hline
\makecell{Duckietown \cite{duckietown}\\ } & \makecell{SS \\ } & \makecell{Low \\ } & \makecell{\cmark \\ } & \makecell{Education \\ } & \makecell{Robotics \\ fundamentals} & \makecell{Basic \\ Sensors} & \makecell{Entry- \\ Level} & \makecell{Extensive \\ } & \makecell{Customizable \\ } & \makecell{\cmark \\ } & \makecell{\xmark \\ } \\
\hline
\makecell{NVIDIA \\ JetRacer \cite{jetracerweb}} & \makecell{SS \\ } & \makecell{Low \\ } & \makecell{\cmark \\ } & \makecell{Hobbyists, \\ Academic} & \makecell{Introductory \\ AI} & \makecell{Cameras \\ } & \makecell{Low- \\ Speed} & \makecell{Moderate \\ } & \makecell{Customizable \\ } & \makecell{\cmark \\ } & \makecell{\xmark \\ } \\
\hline
\makecell{Donkey Car \cite{donkeycar}\\ } & \makecell{SS \\ } & \makecell{Low \\ } & \makecell{\cmark \\ } & \makecell{Hobbyists \\ } & \makecell{Autonomous \\ experimentation} & \makecell{Camera \\ IMU} & \makecell{Entry- \\ Level} & \makecell{Moderate \\ } & \makecell{Customizable \\ } & \makecell{\cmark \\ } & \makecell{\xmark \\ } \\
\hline
\makecell{RoboRacer \\ (F1TENTH) \cite{f1tenth}} & \makecell{SS \\ } & \makecell{Med. \\ } & \makecell{\cmark \\ } & \makecell{Academic, \\ Research} & \makecell{Autonomous racing, \\ algorithm testing} & \makecell{Cameras, LiDAR, \\ IMU} & \makecell{Medium- \\ Speed} & \makecell{Extensive \\ } & \makecell{Customizable \\ } & \makecell{\cmark \\ } & \makecell{\cmark \\ } \\
\hline
\makecell{Abu Dhabi Autonomous \\ Racing League \cite{a2rl}} & \makecell{FS \\ } & \makecell{High \\ } & \makecell{\xmark \\ } & \makecell{Industry, \\ Academic} & \makecell{High-performance \\ autonomous racing} & \makecell{Cameras, \\ LiDAR} & \makecell{High- \\ Speed} & \makecell{Moderate \\ } & \makecell{Proprietary \\ } & \makecell{\xmark \\ } & \makecell{\xmark \\ } \\
\hline
\makecell{Formula Student \\ Driverless (FSD) \cite{fsdweb}} & \makecell{FS \\ } & \makecell{High \\ } & \makecell{\xmark \\ } & \makecell{Academic \\ } & \makecell{High-performance \\ autonomous racing} & \makecell{Cameras, LiDAR, \\ GPS} & \makecell{High- \\ Speed} & \makecell{Extensive \\ } & \makecell{Customizable \\ } & \makecell{\cmark \\ } & \makecell{\xmark \\ } \\
\hline

\multicolumn{12}{l}{OS = Open Source, Comm. Supp. = Community Support, HW Flex. = Hardware Flexibility,  Act. = Active, Sim. = Simulator Available, FS = Full Scale, SS = Small Scale, Med. = Medium } \\

\end{tabular}
}
\label{RACING PLATFORMS}
\end{table*}

The competitive nature of autonomous racing—where vehicles are required to interact with one another, overtake, and optimize lap times—has fostered research into multi-agent systems and competitive learning strategies. In many cases, vehicles must plan and execute complex overtaking maneuvers, requiring tight integration of path planning and control algorithms \cite{10757928}. This has led to advances in areas such as reinforcement learning and multi-agent competitive dynamics, where vehicles learn to compete and adapt to other agents' behaviors in high-speed environments \cite{learningbased}.

Furthermore, RoboRacer competitions and academic research on autonomous racing have broader implications for the development of advanced driver-assistance systems (ADAS) and fully autonomous vehicles. Many techniques honed in racing, such as collision avoidance, sensor fusion, and robust control under uncertainty, directly translate to practical applications in everyday road environments.

\subsection{Autonomous Racing Competitions}

The field of autonomous racing and vehicle research has seen the development of numerous competitions, each designed to cater to specific research objectives and use cases. These competitions, ranging from full-scale systems like the Indy Autonomous Challenge to smaller-scale setups like RoboRacer and Duckietown, provide researchers and developers with diverse environments to test algorithms, sensor integration, and autonomous control strategies. Each competitions offers unique strengths, such as scalability, affordability, or advanced hardware capabilities, making them invaluable tools for academic research, industry innovation, and education.

Table~\ref{RACING PLATFORMS} summarizes the key features, capabilities, and applications of prominent autonomous racing platforms. Among these, RoboRacer stands out for its modular design, extensive community support, and suitability for both academic research and competitive environments. Its ability to replicate real-world challenges at a small scale while maintaining affordability makes it an exceptional platform for advancing autonomous driving technologies. The table highlights how each platform contributes to the field, offering insights into their diverse applications and strengths.

\section{Overview of the RoboRacer Platform}

The RoboRacer platform is designed to advance research and education in autonomous driving and robotics. Guided by principles of scalability, modularity, and self-sufficiency, it offers a realistic and adaptable environment for testing, development, and innovation. Its ability to replicate real-world vehicle dynamics, integrate seamlessly with simulation tools, and foster collaborative experimentation has established it as a cornerstone of autonomous systems research.

\subsection{Hardware and Software Architecture}

The platform combines robust hardware, such as the Traxxas Slash 4x4 chassis, with advanced sensors like 2D LiDAR, cameras (e.g., Intel RealSense), and an IMU for perception and navigation. An onboard NVIDIA Jetson computer processes real-time data for tasks like obstacle detection, SLAM, and control \cite{f1tenthbuild}. 

The software framework leverages the Robot Operating System (ROS), enabling seamless communication between hardware and high-level modules for perception, planning, and control. Algorithms such as Pure Pursuit, Model Predictive Control (MPC), and reinforcement learning can be easily integrated and tested using the platform's flexible and modular architecture \cite{f1tenthros}.

\subsection{Modularity and Customization}

The platform's modular design allows researchers to:
\begin{itemize}
    \item Replace or upgrade sensors and computational units.
    \item Implement and test diverse algorithms without extensive reconfiguration.
    \item Operate independently with onboard power and computational resources.
\end{itemize}
This flexibility ensures the platform remains relevant for cutting-edge research in dynamic and diverse environments.

\subsection{Educational and Research Applications}

RoboRacer serves as a valuable educational tool, offering hands-on opportunities for students to build, program, and race autonomous vehicles. Courses centered around the platform help bridge theoretical knowledge with practical implementation, fostering skills in robotics, control systems, and algorithm development \cite{f1tentheducation}.

In research, the platform supports a wide range of studies, from classical control approaches to advanced learning-based methods. Its adaptability enables experiments in areas such as trajectory planning, collision avoidance, and multi-agent coordination, contributing to broader applications including Advanced Driver Assistance Systems (ADAS) \cite{unifyingf1tenthautonomousracing}.

\subsection{Advancements in Autonomous Racing and Real-World Integration}

The RoboRacer platform plays a pivotal role in advancing autonomous racing by testing algorithms in high-speed, dynamic environments. Researchers have investigated control strategies such as MPC, reinforcement learning, and sensor fusion to enhance decision-making and trajectory planning \cite{unifyingf1tenthautonomousracing}.

Integration with simulation environments like Gazebo enables safe and cost-effective testing prior to real-world deployment. These simulators provide realistic scenarios for validating algorithms, facilitating smooth transitions from virtual environments to physical vehicles \cite{buf1tenth}.

\subsection{Community and Competitions}

The RoboRacer (F1TENTH) community actively engages in global competitions, driving innovation in autonomous racing. These events provide a standardized platform for testing software and algorithms, fostering collaboration and advancing the field. The vibrant open-source community encourages shared knowledge and continuous improvement \cite{ucff1tenth}.

\begin{table*}[!t]
    \caption{Comparison of Simulation Environments for RoboRacer (F1TENTH)}
    \centering
    \setlength{\tabcolsep}{4.5pt}
    \begin{tabular}{|c||c|c|c|c|c|c|c|c|c|c|c|}
    \hline
    \multirow{2}{*}{{\textbf{\textit{Simulator}}}} & 
    \multirow{2}{*}{{\textbf{\textit{Key Features}}{\textit{}}}} & 
    \multirow{2}{*}{\rotatebox{-90}{\parbox{1.5cm}{\textbf{\textit{ROS}}\\\textbf{\textit{COMP}}}}} & 
    \multicolumn{4}{|c|}{\multirow{2}{*}{\textbf{\textit{Sensor Suite}}}}& 
    \multirow{2}{*}{\rotatebox{-90}{\parbox{2cm}{\textbf{\textit{Photorealistic}}\\\textbf{\textit{Rendering}}}}} & 
    \rotatebox{-90}{\parbox{2cm}{\textbf{\textit{Multi-Agent}}\\\textbf{\textit{Support}}}} & 
    \rotatebox{-90}{\parbox{2cm}{\textbf{\textit{Platform}}\\\textbf{\textit{Type}}}} & 
    {\multirow{2}{*}{\textbf{\textit{Applications}}}} \\
    \cline{4-7}
    & & & \rotatebox{-90}{\textit{LiDAR}} & \rotatebox{-90}{\textit{Camera\:}} & \rotatebox{-90}{\textit{GNSS}} & \rotatebox{-90}{\textit{IMU}} & & & & \\
    \hline
    F1TENTH Gym \cite{opensourceevalutaion}, \cite{f1tenthgym} & Deterministic physics & No & \cmark & \xmark & \xmark & \xmark & No & Yes & 2D & Rapid prototyping \\
    \hline
    F1TENTH Gym ROS \cite{f1tenthgymros} & Containerized setup & Yes & \cmark & \xmark & \xmark & \xmark & No & Yes & 2D & Collaborative research \\
    \hline
    AutoDRIVE \cite{autodrivepaper1}, \cite{autodriveweb} & Modular environments & Yes & \cmark & \cmark & \xmark & \cmark & Yes & Yes & 3D & Cooperative multi-agent RL \\
    \hline
    CARLA \cite{carlapaper}, \cite{carlaweb} & Scalable multi-client architecture & Yes & \cmark & \cmark & \cmark & \xmark & Yes & Yes & 3D & Complex driving scenarios \\
    \hline
    AWSIM \cite{awsim}, \cite{awsimautowareweb} & Photorealistic digital twins & Yes & \cmark & \cmark & \cmark & \cmark & Yes & No & 3D & Perception, Planning, and Control \\
    \hline
    Gazebo \cite{gazebo} & Sensor emulation & Yes  & \cmark & \cmark & \xmark & \xmark & No & No & 3D & Path planning, Vehicle control \\
    \hline
    SVL Simulator \cite{lgsvlpaper} & Modular design, Sensor emulation & Yes  & \cmark & \cmark & \cmark & \xmark & Yes & No & 3D & Autonomous racing \\
    \hline
    Learn-to-Race \cite{l2rpaper1}, \cite{l2rpaper2} & High-fidelity racing maps, RL support & No  & \cmark & \cmark & \xmark & \xmark & Yes & Yes & 3D & Autonomous racing \\
    \hline
    \end{tabular}
    \label{tab:simulator_comparison}
\end{table*}

\section{Simulation Environments}
Simulation environments are an integral part of developing, testing, and refining autonomous driving algorithms for the RoboRacer platform. They provide a controlled, cost-effective, and risk-free setting to experiment with perception, planning, and control strategies prior to deployment on physical vehicles. By replicating real-world conditions in a virtual space, simulations enable researchers and developers to evaluate their algorithms across diverse scenarios, including high-speed racing and multi-agent interactions.

The RoboRacer community leverages a range of simulators tailored to address the unique challenges of autonomous racing. These simulators vary in fidelity and functionality, offering capabilities such as photorealistic environments, detailed vehicle dynamics, and integration with middleware frameworks like the Robot Operating System (ROS). Ranging from lightweight 2D tools for rapid prototyping to high-fidelity 3D environments supporting reinforcement learning and multi-agent competition, these platforms enable robust validation of autonomous systems.

Table~\ref{tab:simulator_comparison} summarizes the key features, compatibility, and applications of the most prominent simulators, underscoring their critical role in advancing research and innovation within the RoboRacer ecosystem.

\section{Sim2Real: Bridging the Simulation-to-Reality Gap}

The simulation-to-reality (Sim2Real) gap represents the performance discrepancies encountered when transferring algorithms from simulated environments to real-world applications. This challenge is critical in autonomous racing research, where high precision and speed amplify the effects of these discrepancies. Artificial intelligence plays an increasingly vital role in enabling robust transfer from simulation to reality \cite{9544201}, \cite{7125356}.
While simulators like Gazebo and F1TENTH Gym provide controlled, repeatable environments for algorithm testing, they often fail to capture the full complexity of real-world conditions, such as sensor noise, tire-road interactions, and environmental dynamics, leading to performance mismatches.

\subsection{Sim2Real Gap Challenges}

Simulators simplify physics, sensor behavior, and environmental conditions to maintain computational efficiency, which can hinder real-world transferability. Differences between simulated and actual sensor data, such as noise and response dynamics, can result in perception errors. Additionally, algorithms, especially deep reinforcement learning (DRL) models, may overfit to simulation constraints, reducing their ability to generalize to real-world scenarios. Simulated environments often lack critical real-world variables like dynamic lighting, weather changes, and interactions with other agents \cite{sim2gappaper1}.

\subsection{Strategies to Bridge the Sim2Real Gap}

To address these challenges, researchers employ strategies such as domain randomization, domain adaptation, and iterative real-world testing:

\begin{itemize}
    \item \textbf{Domain Randomization:} Introduces variations in simulation parameters (e.g., textures, lighting) during training to improve generalization to real-world conditions \cite{sim2realdomainrand}.
    \item \textbf{Domain Adaptation:} Fine-tunes models using real-world data to reduce discrepancies between simulated and real environments, particularly for perception modules \cite{domainadaptation}.
    \item \textbf{Iterative Testing:} Combines simulation and real-world validation, allowing for continuous refinement of algorithms to address discrepancies and unexpected conditions.
\end{itemize}

The RoboRacer platform supports seamless transitions between simulation and reality through its modular design and integrated ecosystem. Researchers can develop, test, and refine algorithms iteratively, using domain randomization and adaptation techniques to improve robustness. This approach ensures that models trained in simulation can reliably operate under real-world racing conditions, effectively narrowing the Sim2Real gap.

By fostering an open-source, collaborative environment, the RoboRacer platform advances autonomous racing research, making it a valuable tool for addressing the complexities of Sim2Real transfer.

\section{Research Resources Supporting the RoboRacer Platform}

Datasets are critical for developing and evaluating autonomous vehicle algorithms in both simulated and real-world settings \cite{10385062}. Within the RoboRacer research community, datasets covering perception, planning, and control tasks serve as benchmarks for algorithmic performance. Below is an overview of key resources and their significance for advancing autonomous racing.

\subsection{Maps and LiDAR Datasets}
The RoboRacer platform offers a variety of track maps, including standard competition circuits and community-generated layouts \cite{maps1}, \cite{maps2}. These maps support diverse testing scenarios for perception and planning strategies, Fig. 1. Complementary LiDAR datasets introduced by Zarrar et al. \cite{lidargit} and paired with TinyLidarNet \cite{tinylidarpaper} enable lightweight, 2D LiDAR-based research, advancing end-to-end deep learning models for racing.

\begin{figure*}[!t]
  \centering  \includegraphics[width=1\linewidth]{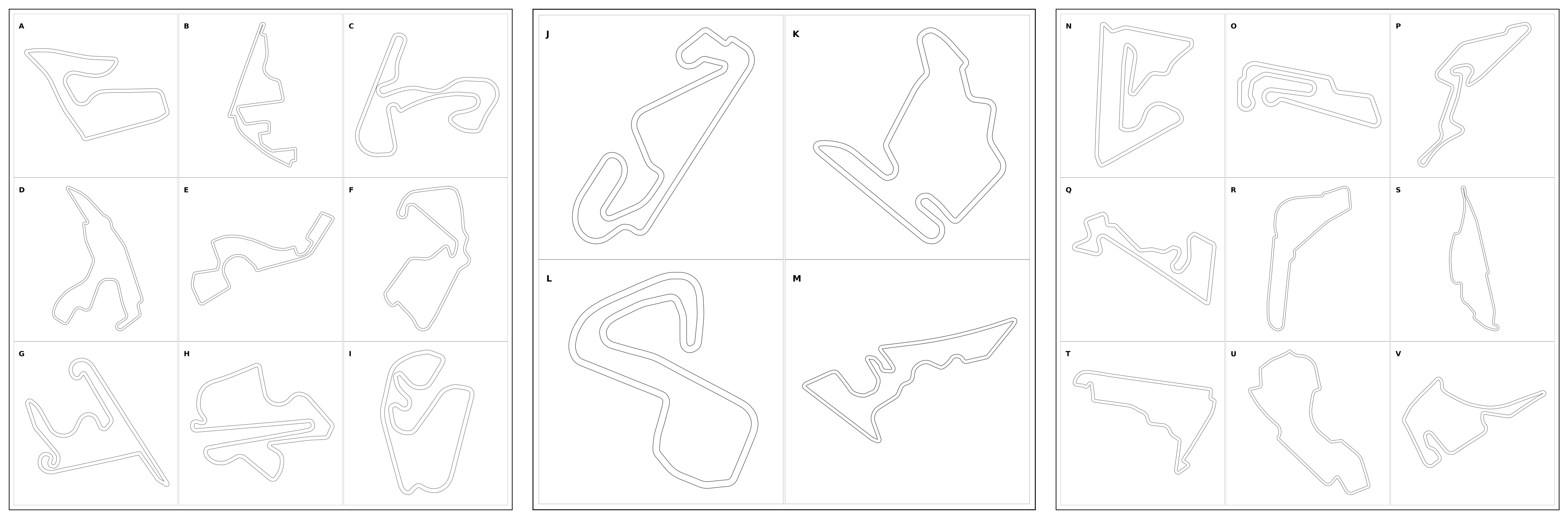}
  \vspace{-.5cm}
  \caption{Collage of F1Tenth (RoboRacer) Track Maps. Locations:  (A) Spielberg, Austria; (B) Yas Marina, Abu Dhabi, UAE; (C) Zandvoort, Netherlands; (D) Spa, Belgium; (E) Sochi, Russia; (F) Silverstone, UK; (G) Shanghai, China; (H) Sepang, Malaysia; (I) São Paulo, Brazil; (J) Catalunya, Spain; (K) Budapest, Hungary; (L) Brands Hatch, UK; (M) Austin, USA; (N) Sakhir, Bahrain; (O) Oschersleben, Germany; (P) Nürburgring, Germany; (Q) Moscow Raceway, Russia; (R) Monza, Italy; (S) Montreal, Canada; (T) Mexico City, Mexico; (U) Melbourne, Australia; (V) Hockenheim, Germany.  
  \textit{Note: The track widths were extracted from satellite imagery using an image processing algorithm. The data was then downscaled to a 1:10 scale, with a fixed track width of 2.20 m, to ensure compatibility with RoboRacer (F1Tenth) environments.}}

  \label{Map}
\end{figure*}

\subsection{Vision-Based Datasets}
Vision datasets curated by the community include labeled resources for training object detection models using YOLO and PyTorch \cite{cnndatasets}. Additionally, pretrained models hosted on Roboflow Universe \cite{roboflowtop}, \cite{roboflow} facilitate object detection and classification, providing a quick start for vision-based research \cite{7395239}, \cite{heidarizadeh2021preprocessing}, \cite{7125330}.

\subsection{Trajectory Data and Global Planning}
Trajectory datasets play a crucial role in studying vehicle behavior, path optimization, and interaction modeling. One such dataset includes over 300 recorded trajectories from 30 agents across 251 timesteps, offering valuable insight into multi-agent dynamics and planning under racing conditions \cite{trajectories1}, \cite{trajectories2}. 

In addition, global planning tools developed by the Technical University of Munich enhance route optimization by incorporating detailed friction maps and precise track geometry, supporting research in physics-informed planning and trajectory refinement \cite{racetrackoptimization}.

\subsection{Common RoboRacer Algorithms}
Standardized implementations of RoboRacer algorithms—including classical controllers, reinforcement learning strategies, and perception modules—are publicly available \cite{unifyingf1tenthautonomousracing}, \cite{f1tenthbenchmarks}. These benchmarks provide reliable baselines for evaluating new methods against state-of-the-art solutions.

\subsection{Significance of Datasets}
The shared availability of these datasets fosters collaboration and accelerates innovation in autonomous racing:
\begin{itemize}
    \item \textbf{Benchmarking Performance:} Datasets enable standardized comparisons using metrics like lap time, trajectory fidelity, and collision rates.
    \item \textbf{Validating Algorithms:} Real-world and simulated data allow researchers to refine and verify algorithms before deploying them on hardware.
    \item \textbf{Encouraging Collaboration:} Open-source datasets promote shared progress, reproducibility, and collective advancements in high-speed autonomy.
\end{itemize}

Together, these datasets underscore the depth of the RoboRacer research ecosystem, ensuring reproducibility and transparency while driving innovation in high-speed autonomy.

\section{Control Algorithms for RoboRacer}
Control algorithms form the backbone of the RoboRacer platform, translating high-level planning and perception data into low-level commands for steering, speed, and trajectory control. Their effectiveness directly impacts performance, stability, and safety in dynamic and competitive racing environments.

This section highlights key control strategies employed on the platform, emphasizing their relevance, functionality, and practical applications.

\subsection{Traditional Control Algorithms}

\textbf{PID Controller:} Widely used for wall-following and low-level control, PID controllers adjust steering and speed based on the error between a desired setpoint and the actual state. While simple and effective, their performance depends on precise tuning of proportional, integral, and derivative gains. Adaptive PID systems address this limitation by dynamically adjusting parameters in response to real-time conditions \cite{pidwallfollowing}.

\textbf{Pure Pursuit:} A geometric path-tracking algorithm that computes steering angles to align the vehicle with a look-ahead point on a predefined path. Its simplicity and effectiveness make it suitable for waypoint following, although adaptive variants are often required to handle varying speeds and track curvatures \cite{purepursuitlab}.

\textbf{Stanley Controller:} Known for precise lateral control, the Stanley controller minimizes cross-track and heading errors to align the vehicle with a desired trajectory. Its robustness and adaptability have made it a popular choice for autonomous path tracking \cite{stanley}.

\subsection{Learning-Based Methods}

\textbf{Reinforcement Learning (RL):} RL techniques such as Proximal Policy Optimization (PPO) enable vehicles to adapt to diverse racing scenarios by learning optimal behaviors through interaction with simulation environments. These methods excel in handling dynamic and unpredictable conditions, though challenges such as data efficiency and the Sim2Real gap remain \cite{rlf1tenth}.

\textbf{Recurrent Neural Networks (RNNs) and LSTMs:} These data-driven models predict vehicle behavior over time by capturing temporal dependencies for more responsive control. LSTMs, with their capacity to manage long-term dependencies, have shown promise in modeling complex vehicle dynamics \cite{learningbased}.

\subsection{Model Predictive Control (MPC)}
MPC leverages a dynamic vehicle model to predict future states and optimize control inputs over a defined time horizon. By balancing performance objectives and system constraints, it enables safe and efficient trajectory planning. Although computationally intensive, MPC is a powerful method for multivariable control in dynamic environments \cite{mpcf1tenth}.

\subsection{Reactive Methods: Follow-The-Gap (FTG)}
FTG is a computationally efficient obstacle avoidance algorithm that dynamically steers the vehicle toward the largest navigable gap in sensor data. It allows autonomous vehicles to navigate complex environments in real time without requiring predefined maps. Its simplicity and speed make it ideal for dynamic, real-world applications \cite{followthegaplab}.

\subsection{Summary of Control Strategies}

Table~\ref{tab:control_algorithms} provides an overview of the key control strategies implemented on the RoboRacer platform, highlighting their core principles, applications, and strengths in addressing the challenges of high-speed autonomous navigation.

\begin{table}[t!]
\centering
\caption{Key Control Algorithms on RoboRacer (F1TENTH)}
\label{tab:control_algorithms}
\small
\begin{tabular}{
    |>{\centering\arraybackslash}p{1.3cm}||
     >{\centering\arraybackslash}p{3.6cm}|
     >{\centering\arraybackslash}p{2.6cm}|}
\hline
\textbf{Algorithm} & \textbf{Key Features} & \textbf{Applications} \\ \hline
PID & Simple, requires precise tuning, adaptive versions & Wall-following, low-level control \\ \hline
Pure Pursuit & Geometric path tracking, look-ahead distance critical & \footnotesize Waypoint following, \small smooth trajectory \\ \hline
Stanley & Minimizes cross-track and heading error & Precise path tracking, stable behavior \\ \hline
MPC &  Predictive, computationally demanding, handles constraints & Trajectory optimization \\ \hline
FTG & Reactive, avoids predefined maps, efficient & Real-time obstacle avoidance \\ \hline
RL & Learns from interaction, adapts to diverse scenarios & Dynamic racing, complex setups \\ \hline
LSTMs & Captures long-term temporal dependencies & Predictive nonlinear control \\ \hline
\end{tabular}
\end{table}

\section{RoboRacer Competitions}

The RoboRacer platform has cultivated a vibrant community through competitions that challenge participants to design, build, and program autonomous race cars. These events provide a practical arena for advancing autonomous vehicle technologies and fostering collaboration among researchers, engineers, and enthusiasts.

RoboRacer competitions are featured at several prestigious conferences, offering a platform to showcase advancements in autonomous systems and intelligent vehicle technologies. Table~\ref{tab:f1tenth_competitions} summarizes key conferences hosting these events.

These conferences and competitions play a pivotal role in fostering innovation and collaboration. By providing platforms for hands-on experimentation and showcasing state-of-the-art solutions, RoboRacer competitions continue to drive advancements in autonomous vehicle research and development.

\begin{table}[t!]
\centering
\caption{Key Conferences Hosting RoboRacer (F1TENTH) Competitions}
\label{tab:f1tenth_competitions}
\renewcommand{\arraystretch}{1.5} 
\small
\begin{tabular}{
    |>{\centering\arraybackslash}m{1.4cm}||
     >{\centering\arraybackslash}m{1.0cm}|
     >{\centering\arraybackslash}m{1.5cm}|
     >{\centering\arraybackslash}m{3.1cm}|}
\hline
\textbf{Conference} & \makecell{\textbf{Year}\\\textbf{Edition}} & \textbf{Location} & \textbf{Highlights} \\ \hline
ICRA \cite{icra2024} & 2024 (15th) & Yokohama, Japan & Premier robotics conference showcasing autonomous racing. \\ \hline
CDC \cite{cdc2024} & 2024 (22nd) & Milan, Italy & Features advanced control strategies for autonomous racing. \\ \hline
ITSC \cite{itsc2024} & 2024 (20th) & Edmonton, Canada & Emphasizes intelligent transportation solutions. \\ \hline
IV \cite{iv2024} & 2024 (18th) & Jeju Island, Korea & Focuses on intelligent vehicle technologies. \\ \hline
SM \cite{sm2024} & 2024 (19th) & Ontario, Canada & Highlights smart mobility innovations. \\ \hline
IROS \cite{iros2024} & 2024 (21st) & Abu Dhabi, UAE & Global platform for cutting-edge robotics systems. \\ \hline
CPS-IoT Week \cite{cpsweek2024} & 2024 (17th) & Hong Kong & Real-time and embedded systems with racing challenges. \\ \hline
\end{tabular}
\end{table}

\section{Conclusion}
This paper presented a comprehensive review of the RoboRacer (F1TENTH) platform, emphasizing its contributions to advancing autonomous driving research and education. By providing a modular, scalable, and cost-effective testbed, RoboRacer (F1TENTH) enables rigorous development and testing of perception, planning, and control algorithms across both simulated and real-world scenarios. The platform's open-source framework, standardized datasets, and global competitions foster collaboration, innovation, and benchmarking within the community. Furthermore, its seamless integration of simulation and real-world testing supports bridging the Sim2Real gap, addressing critical challenges in autonomous racing and beyond. As autonomous systems continue to evolve, RoboRacer (F1TENTH) remains a vital resource for pushing the boundaries of robotics and intelligent transportation technologies.

\bibliographystyle{IEEEbib}
\bibliography{references}

\end{document}